\documentclass[runningheads,a4paper]{llncs}

\usepackage{amssymb}
\usepackage{graphicx}
\usepackage{tikz}

\usepackage{url}
\urldef{\mailsa}\path|{alfred.hofmann, ursula.barth, ingrid.haas, frank.holzwarth,|
\urldef{\mailsb}\path|anna.kramer, leonie.kunz, christine.reiss, nicole.sator,|
\urldef{\mailsc}\path|erika.siebert-cole, peter.strasser, lncs}@springer.com|    
\newcommand{\keywords}[1]{\par\addvspace\baselineskip
\noindent\keywordname\enspace\ignorespaces#1}

\begin{document}

\mainmatter  

\title{Understanding Humans' Strategies in Maze Solving}

\titlerunning{Understanding Humans' Strategies in Maze Solving}

%
%
\author{Min Zhao, Andre G. Marquez}

%
\authorrunning{Understanding Humans' Strategies in Maze Solving}

\institute{Department of Psychology, Rutgers University-New Brunswick\\
152 Frelinghuysen Road, Piscataway, NJ 08854, USA\\minzhao@rci.rutgers.edu}

%
%

\toctitle{Understanding Humans' Strategies in Maze Solving}
\tocauthor{Dept. of Psychology, Rutgers University}
\maketitle

\begin{abstract}
Navigating through a visual maze relies on the strategic use of eye movements to select and identify the route.  When navigating the maze, there are trade-offs between exploring to the environment and relying on memory. This study examined strategies used to navigating through novel and familiar mazes that were viewed from above and traversed by a mouse cursor. Eye and mouse movements revealed two modes that almost never occurred concurrently:  exploration and guidance.  Analyses showed that people learned mazes and were able to devise and carry out complex, multi-faceted strategies that traded-off visual exploration against active motor performance. These strategies took into account available visual information, memory, confidence, the estimated cost in time for exploration, and idiosyncratic tolerance for error. Understanding the strategies humans used for maze solving is valuable for applications in cognitive neuroscience as well as in AI, robotics and human-robot interactions. 
\end{abstract}

\section{Introduction}

\subsection{Maze solving}

Maze solving is an important topic within artificial intelligence. Many algorithms have been developed to solve mazes, such as the Random Mouse algorithm, Wall Follower, Pledge algorithm and Dead-end filling (Even, 2011; Sedgewick, 2002). These algorithms either treat the agent in the maze as unintelligent, which means it uses sensors and does not have memory, or as partially intelligent, which means it uses sensors, but also remembers and uses the previous states. These approaches show that when an agent solves a maze, there is a trade-off between the use of sensors and the number of states to be remembered. 

A human being solving a maze also relies on sensors and on memory to decide which path to take. This is evident when solving an overhead maze with the hand or a mouse. In order to solve the maze quickly, people need to trade-off scanning ahead with the eye for more information, and so relying on memory to store the correct path, vs. a strong reliance on immediate visual cues, at the risk of time-consuming errors. When trading-off scanning ahead for information vs. immediate visual guidance, people may take into account memory load, as well as the precision of the visual information at each scanning spot. An outcome of this trade-off determines the strategy people choose. Studying eye-hand coordination during maze solving should be a good way to determine the strategy. Understanding the strategies humans use for maze-solving is valuable for applications in cognitive neuroscience, such as understanding the mechanisms that determine the optimal use of resource in natural tasks, as well as applications in AI or robotics, where a better understanding of human strategies could be helpful to inform models that guide robots or guide human-robot interactions.

\subsection{Eye-Hand Coordination}

Eye movements can be recorded precisely in both spatial and temporal domains. A crucial question remains: what can eye movements tell us about underlying cognitive processes, such as memory, reasoning, or planning. Viviani (1990) pointed out that to understand cognitive process from eye movements, one should build up a theoretical framework in which cognitive processes are sequentially linked to the sequence of eye movements. Recent studies of eye-hand coordination have added a new dimension (hand movements) as a way to unearth the underlying cognitive processes. It is well known that there is a collaborative pattern between eye and hand movements, in that the eye usually searches for the target ahead and then guides the hand to the target (e.g. Ballard, et. al., 1995; Epelboim, et. al., 1997; Flanagan, et. al., 2003). Those prior studies also showed that a major motivation guiding strategies is to reduce reliance on memory.

\subsection{Current Study}

By tracking both eye and hand movements during solving of overhead mazes, the current study aims to reveal the strategies, as well as the underlying cognitive processes, that determines perceptual-motor cooperation. The present study also examined learning. Learning was investigated by presenting mazes twice in different relative spatial configurations, in order to determine how much was learned, and how learning influenced strategies of maze solving.

\section{Method}

\subsection{Subjects}

25 subjects were tested, all with approval of the Rutgers University Institutional Review Board for the Protection of human subjects. 23 out of these 25 subjects were undergraduates recruited from the General Psychology subject pool who earned course credits. The other 2 were paid volunteers. All subjects had normal vision. All subjects completed 6 sessions, with 22 trials in each session. 

\subsection{Stimuli}

\begin{figure}
\centering
\includegraphics[width=0.5\textwidth,height = 6cm]{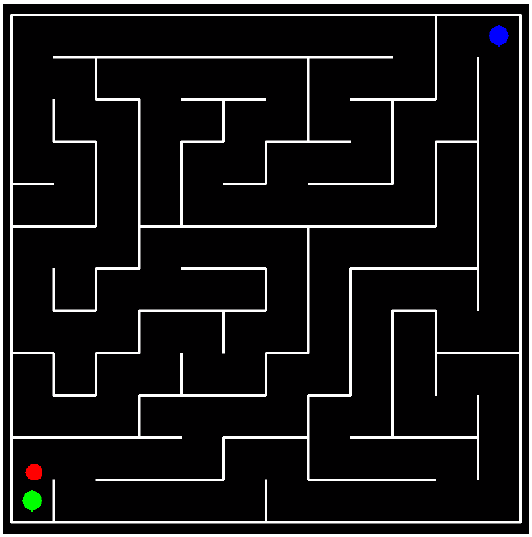}
\caption{An example of a maze used in the experiment.}
\label{fig:maze}
\end{figure}

Stimuli were presented on the Viewsonic G90fb CRT monitor, 1024*786 resolution, 60 Hz refresh rate. The display area subtended 16.28 horizontally by 12.38 vertically and was viewed from a distance of 119 cm. Stimuli were square 12 unit by 12 unit mazes, 10 degrees visual angle on each side, centered on the screen, (Fig.~\ref{fig:maze}). The start location was assigned to one of the four corners of the maze and the end location was always the opposite corner. Mazes were generated by a free random maze generator (written in python language by Georgy Pruss, 2003, http://code.activestate.com/recipes/578356-random-maze-generator/ ). 60 mazes were randomly selected as the stimuli. 

\subsection{Eye movement recording}

Eye movements were recorded using the Eyelink 1000 (SR Research, Osgoode, Canada) tower mounted version, sampling at 1000 Hz. A chin rest was used to stabilize the head. Eye movements were recorded from the right eye. 

\subsection{Procedure}

Each session contained 20 maze trials and a separate calibration trial at the beginning and again at the end. In the two calibration trials, subjects were asked to look at and click on each corner of a square by a mouse. Calibration trials were used to verify the accuracy and precision of the mouse and eye signals, supplementing the usual Eyelink 9-point calibration routine. 

Before every maze trial, subjects were asked to fixate at a central cross and click the mouse to start the trial when ready. After fixating on the cross for 1 second, the start point (green disc) and the end point (blue disc) appeared on the display. Subjects moved the cursor (red disc) to the start location and clicked on the start location. After a 500 ms delay, the maze appeared and subjects could begin to use the mouse to navigate the cursor in the maze from the start location to the end location. Once the cursor reached the end location the trial would end automatically. 

Each maze was solved twice consecutively. The first trial of each pair was the ``training trial'' and the second trial of each pair was the ``testing trial''. 

\begin{figure}
\centering
\includegraphics[width=0.8\textwidth,height = 6cm]{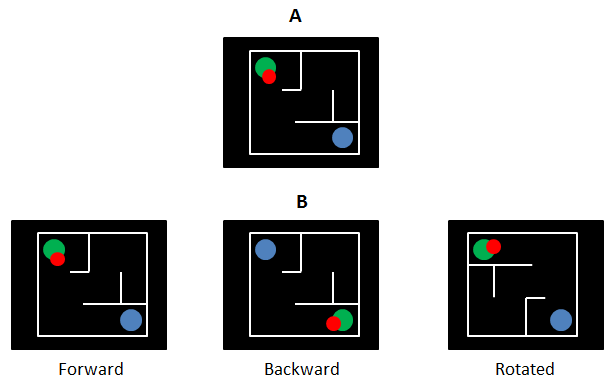}
\caption{Three types of maze spatial relationship between training trials and following testing trials. This is a 4 by 3 maze for illustration purpose. Mazes used in the experiment are much larger.}
\label{fig:spatialrelationship}
\end{figure}

There were three different types of spatial relationships between the training and testing trials: \emph{forward condition}, in which the testing trial used exactly the same maze as the training trial; \emph{backward condition}, in which the testing trial used the same maze as the training trial, but the start and the end locations were exchanged; \emph{rotated condition}, in which the maze in the testing trial was rotated 180 degrees from the maze in the training trial, and the start and the end locations remained at the same positions on the screen. Fig.~\ref{fig:spatialrelationship} illustrates these three conditions, using a small maze (4 by 3) for illustration purposes. 

Subjects were divided into 3 groups. The “Expected” group (n=10) was notified that two successive trials with the same maze would be tested. Subjects in “Un-Expected” group (n=10) were not told that the maze would be repeated twice. The “Previewed” group was similar to the ``Expected'' group. The only difference was that the mazes in training trials were presented for 20 s and subjects could freely scan the maze, but were not allowed to solve the maze by mouse.

\subsection{Analysis}

The beginning and ending positions of saccades were detected offline by means of a computer algorithm employing a velocity criterion to find saccade onset and offset. The value of the criterion was determined empirically for individual observers by examining a large sample of analog recordings of eye positions. 

Mouse positions were recorded in every refresh frame. Mouse signals were filtered at 10 Hz frequency (Flash and Hogan, 1985). 

Data reported are based on the analysis of 6 sessions with 20 maze trials each for all subjects.

\section{Results}
\subsection{Initial solving Time}

\begin{figure}
\centering
\includegraphics[height=5cm]{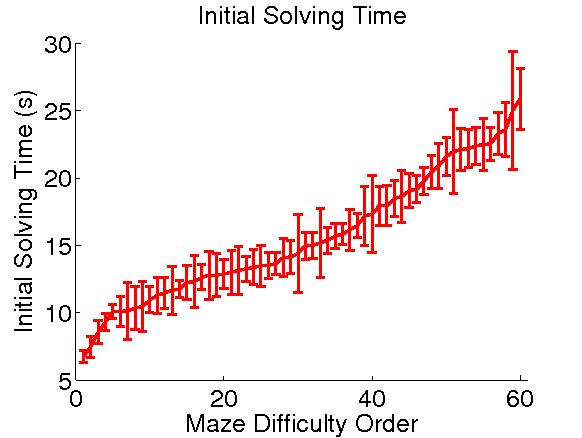}
\caption{Initial Travel time (s). It is sorted from the shortest to longest.}
\label{fig:initialRT}
\end{figure}

The time taken to solve the 60 mazes was determined from the training trials in ``Expected'' and ``Un-Expected'' groups. Average solving time was found for each of the 60 mazes (n = 20 trials/maze). Average solving time (+/- 1 SE) for each maze are shown in order from fastest to slowest in Fig.~\ref{fig:initialRT}. This ordering served as an index of maze difficulty, i.e., the shorter the initial solving time is, the ``easier'' the maze is. 

Initial solving time increased approximated linearly across the 60 mazes. As the initial solving time increased, the variability (individual differences) of subjects’ performance also increased. 

Note that several characteristics accounted for why some mazes were more difficult than others, including total path length, numbers of turns, and the presence in the maze of long ``blind alleys'' that, if entered, required time (back – tracking) to return to the correct path.

\subsection{Analysis of strategies}
\subsubsection{Eye-mouse coordination patterns}

Inspection of recording of eye and mouse movements revealed two modes of performance that almost never occurred concurrently: \emph{exploration}, in which saccades were made to search for the correct path while the mouse was stationary, and \emph{guidance}, in which saccades guided the mouse along the chosen path. 

\begin{figure}
\centering
\includegraphics[width=.8\textwidth,height = 3.5cm]{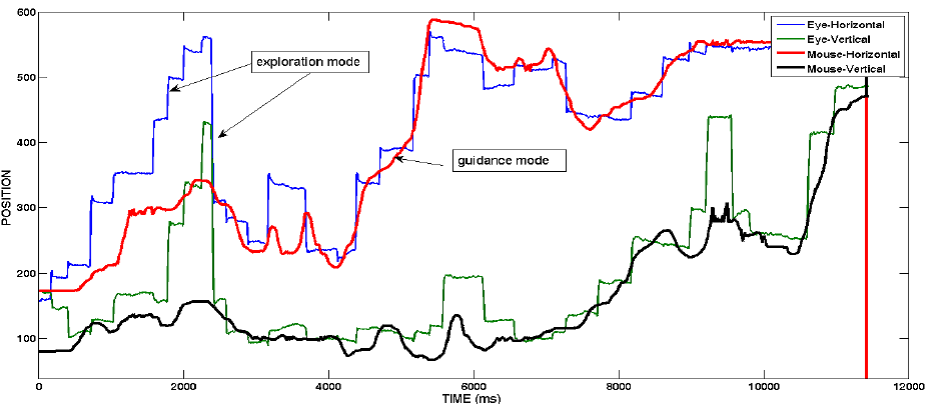}
\caption{Example of eye and mouse traces. Eye positions (blue is horizontal and green is vertical) and mouse positions (red is horizontal and black is vertical) were plotted as a function of time.}
\label{fig:stage1}
\end{figure}

Guidance episodes were highly stereotypical, with the eye executing sequences of saccades along the path and almost always leading the mouse. Exploration was idiosyncratic. Some subjects explored extensively with saccades before beginning to move the mouse. Others alternated between episodes of exploration and guidance. Fig.~\ref{fig:stage1} illustrates these two coordination patterns on the temporal traces of eye and mouse. 

\begin{figure}
\centering
\includegraphics[width=1\textwidth,height = 8cm]{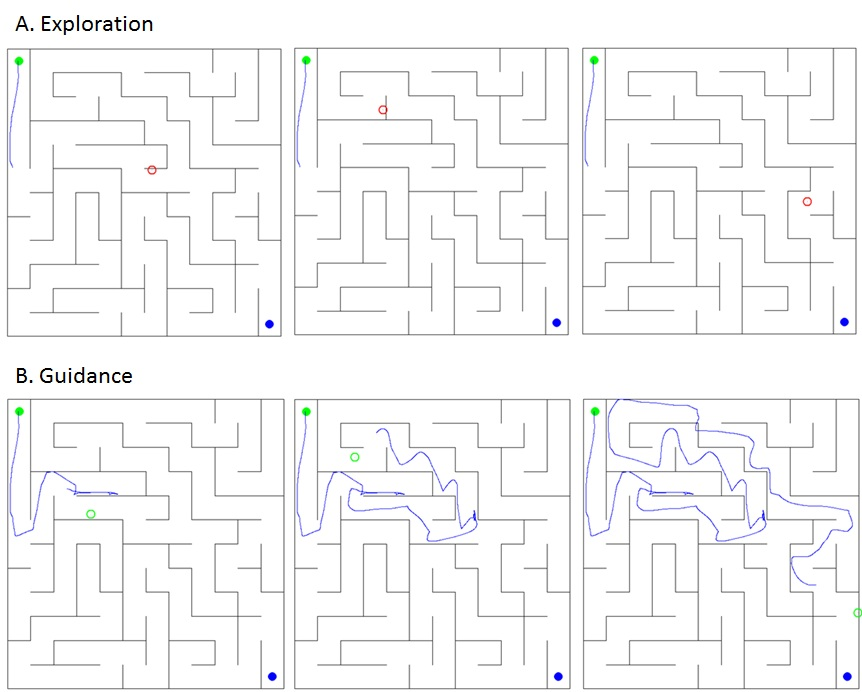}
\caption{Selected frames from dynamic plot of eye and mouse positions in a maze solving trial. There are two main modes of eye-hand coordination (A: Exploration; B. Guidance). Offset of each saccade (circle) and mouse positions (blue line) were plotted on the maze dynamically. There were two patterns: exploration (upper plots) and guidance (lower plots). In exploration mode, mouse (blue line) stopped or slow down and eye (red circle) was sent out to explore in the maze. In guidance mode, eye (green circle) jumped ahead and waited for mouse (blue line) to catch up.}
\label{fig:stage2}
\end{figure}

The differences between exploration and guidance can also be seen when the traces were superimposed on the maze. Fig. 5 shows several frames from a movie of eye and mouse positions on the maze. In exploration mode (Fig.~\ref{fig:stage2}, A), mouse (blue line) slowed down or stopped and the eye (red circle) was sent out to search the maze. In guidance mode (Fig.~\ref{fig:stage2}, B), the eye (green circle) jumped ahead and waited for mouse to catch up.

\begin{figure}
\centering
\includegraphics[width=0.8\textwidth,height = 5cm]{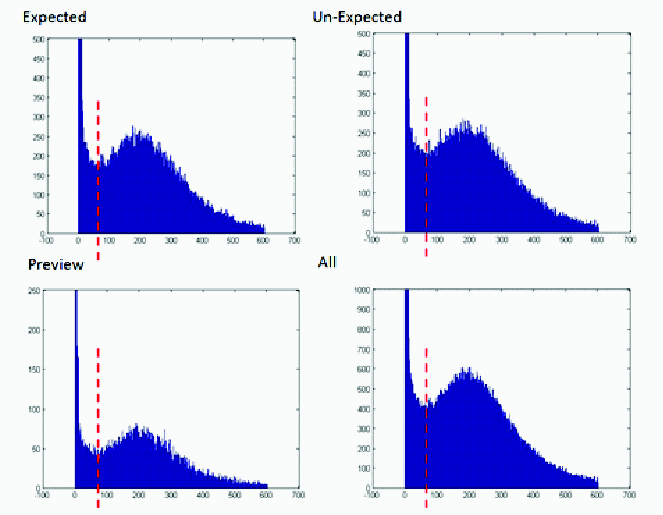}
\caption{Density distribution of mouse speed in each segments in ``Expected'' group (upper left), ``Un-Expected'' group (upper right), Preview group (lower left) and combined all groups (lower right). Criteria (speed = 70 pixel/s, red line) of dividing guidance pattern and exploration pattern based on mouse movements speed.}
\label{fig:criteria}
\end{figure}

Two criteria were used to separate guidance and exploration phases: the speed of mouse movement and the distance between eye and mouse. These criteria were applied to successive segments of eye and mouse movements, where the segments were defined as the time between the onset of one saccade and the onset of the following saccade. For each segment, the speed of mouse movement was calculated (mouse trace length in the segment / duration of the segment). Distributions of speeds are plotted in Fig.~\ref{fig:criteria}. Based on the distributions, 70 pixels/s was selected as the speed criterion: any segment in which the mouse speed was greater than 70 pixels/s was labeled as “guidance”. For the segments in which the mouse speed was less than 70 pixels/s, if the distances between the mouse and eye at the onset and offset of the current saccade were both greater than a distance criteria (set as 50 pixels, which is equals to the width of the path), the current segment was labeled as ``exploration''. A very small proportion of cases, in which the mouse stopped or moved very slowly while the eye was close to the mouse, were assumed to dealing with difficulties of mouse movements, and were categorized as ``guidance''. 

``Guidance'' was segmented further into guiding when the mouse was on the correct path and guiding when the mouse was off the correct path. Thus, there were three phases: exploration; on-path guidance and off-path guidance. The time spent on ``exploration'' was calculated as the sum of the durations of exploration saccade segments. The time spent on ``guiding when mouse on the correct path'' (on-path guidance) was calculated as the sum of the durations of guidance segments when the mouse was traveling on the correct path. The time spent on ``guiding when mouse off the correct path'' (off-path guidance) was calculated as the sum of the durations of guidance saccades segments when the mouse was traveling off the correct path. Most of exploration was conducted when the mouse was on the correct path and only a very small amount of exploration was off-path. Some of explorations were done at the very beginning of the trial before the first movements of the mouse. The sum of time spent on these three phases was the total maze solving time.

\subsubsection{Training trials}

Fig.~\ref{fig:trainingarea} shows how the total maze solving time was apportioned among the 3 phases: on-path guidance, off-path guidance and exploration. Results are shown for the training trials in the ``Expected'' and the ``Un-Expected'' groups as a function of the maze difficulty order (Fig.~\ref{fig:initialRT}). The time spent in each phase is indicated by the width of each band. 

By comparing the width of each band, we can see that the ``on-path guidance'' time was similar between the two groups. The ``on-path guidance'' time increased with maze difficulty, reflecting the longer path and reduced velocity due to many turns in the path. 

The ``off-path guidance'' time also increased with maze difficulty in both groups. However, the increases were small across the mazes tested. Off-path guidance time took up about 10\% to 13\% of the total maze solving time. ``Exploration'' also increased with maze difficulty and accounted for 15\% to 25\% of the solving time. The ``Expected'' group generally spent more time on exploration than the ``Un-Expected'' group. The longer exploration time lead to overall longer solving time in the ``Expected'' group. But it also contributed to decreasing errors, which appeared as about 25\% less time spent ``off-path guidance'' in the ``Expected'' than the ``Un-Expected'' group. Thus, exploration was useful in finding the correct path and avoiding time off the path. 

Although the time spent on ``off-path guidance'' in the ``Expected'' group was less than in the ``Un-Expected'' group, the overall ``off-path guidance'' took up about 16\% of total solving time and the difference was small. The extra time exploring in the ``Expected'' group also added to the total maze solving time. These aspects of results suggest that the main goal of the extra exploration in ``Expected'' group was not to avoid the errors in the current trial, but rather to get and save information about the maze to be used in the next trial, where the same maze would be presented.

\subsubsection{Testing trials}

Fig.~\ref{fig:testingarea} shows how time was apportioned among the three patterns in the testing trials for the ``Expected'' (top row) and ``Un-Expected'' (middle row) groups. The bottom row shows performance for the testing trials in “Preview” group, in which subjects were first presented mazes for 20s in the training trials but not allowed to travel in the maze with the mouse. The columns show the three different spatial relationships between the training maze and the testing maze (see Fig.~\ref{fig:spatialrelationship}): forward (left plots); backward (middle plots); rotated (right plots). 

All three graphs show similar patterns in the testing trial as in training trial, in that the time increased as the maze difficulty increased. However, the time spent in all three phases (on-path, off-path and exploration) depended on the spatial relationship between training and testing mazes. The time spent in the different phases can be seen more clearly in Fig.~\ref{fig:meanbars}, which shows mean time in the three different phases in three separate graphs. 

The time spent in ``on-path guidance'' (green bars) in the testing trials was shortest in the forward condition and longest in the rotated condition. The time on-path in the rotated condition testing trials was actually greater than in the training trials. These results show that learning affected how quickly the maze was traveled, even when on the path. Notice that in the rotated condition, the effect of learning was to slow down the travel time. 

\begin{figure}
\centering
\includegraphics[width=1\textwidth,height=3 cm]{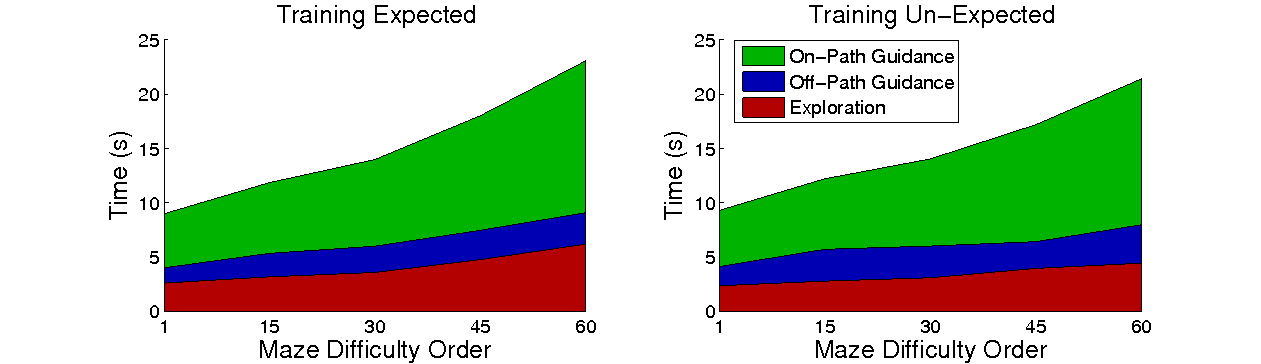}
\caption{Total maze solving time segmented into 3 phases. The time spent in each phase is indicated by the width of each band. Data are shown for training trials as a function of maze difficulty order, from ``“Expected'' group (left plot) and ``Un-Expected'' group (right plot). There were three phases: exploration (red); guidance when the mouse was on the correct path (green); and guidance when the mouse was off the correct path (blue).}
\label{fig:trainingarea}
\end{figure}

\begin{figure}
\centering
\includegraphics[width=1\textwidth,height=10 cm]{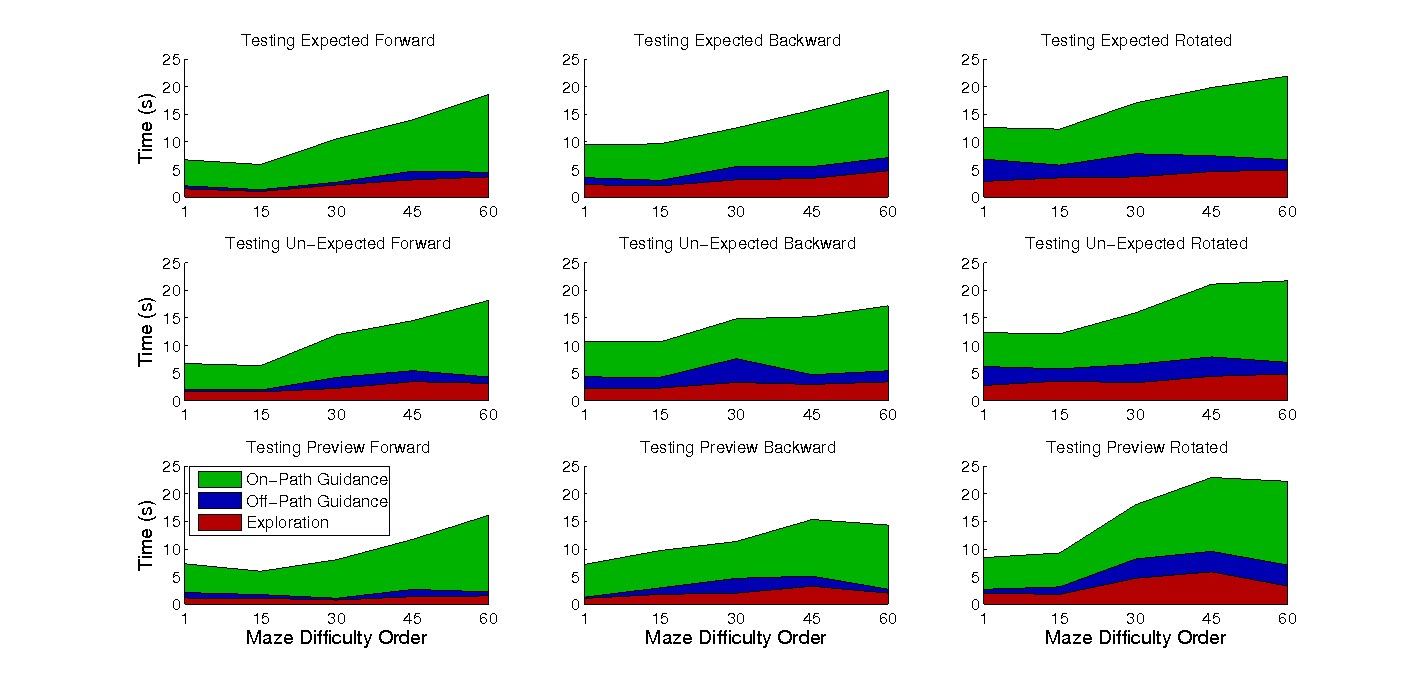}
\caption{Total maze solving the segments into 3 phases. The time spent in each phase is indicated by the width of each band. Data are shown for testing trials as a function of maze difficulty order, from ``Expected'' group (upper plots), ``Un-Expected'' group (middle plots) and Preview group (lower plots). 60 mazes were separated based on the spatial relationships (forward – left; backward – middle; rotated – right). There were three patterns: exploration (red); guidance when the mouse was on the correct path (green); and guidance when the mouse was off the correct path (blue).}
\label{fig:testingarea}
\end{figure}

\begin{figure}
\centering
\includegraphics[width=1\textwidth,height=8 cm]{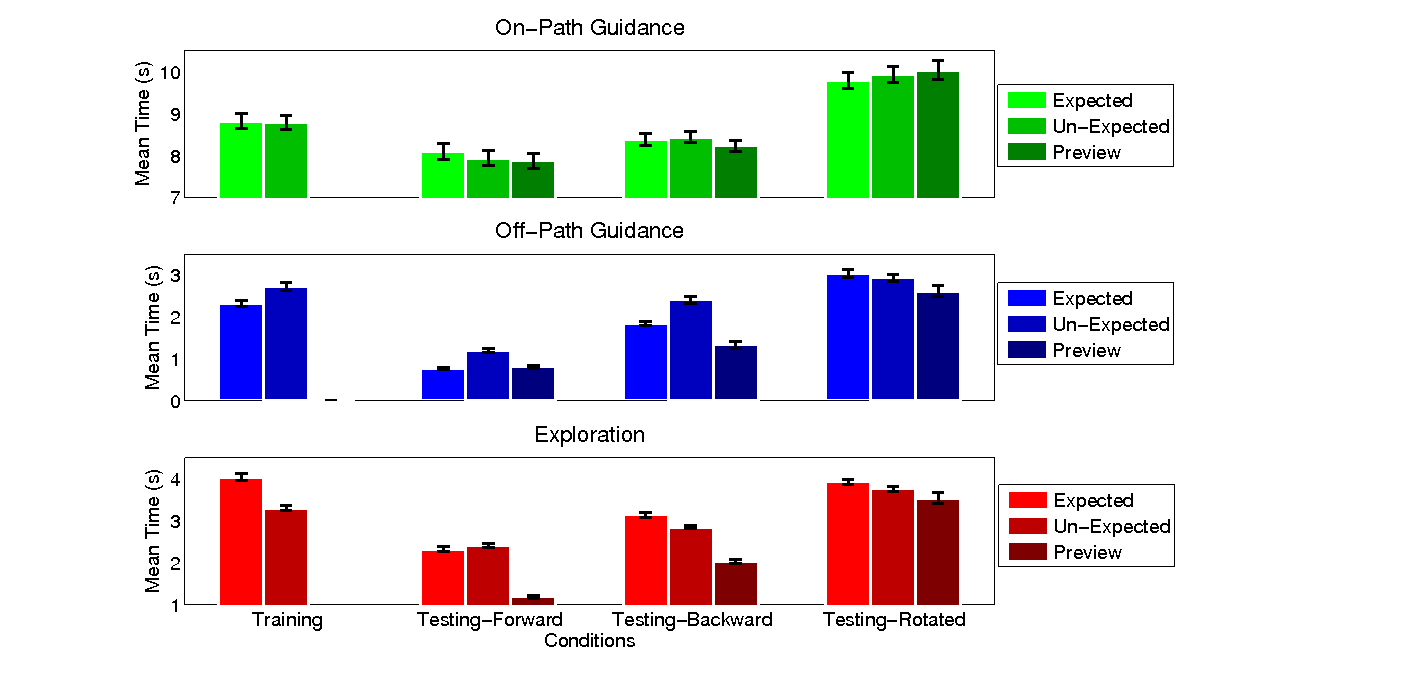}
\caption{Mean times spent in the three different phases of eye-hand coordination patterns: exploration (red, lower plot); guidance when the mouse was on the correct path (green, upper plot); and guidance when the mouse was off the correct path (blue, middle plot). Subjects were from one of three groups:  ``Expected'' group (in lightest color), ``Un-Expected'' group (in the color with mediate brightness) and Preview group (in darkest color). Performance in testing trails was divided into three groups based spatial relationships: forward, backward and rotated.}
\label{fig:meanbars}
\end{figure}

The blue bars (middle) show ``off-path guidance''. This indicates the errors made (travel down the wrong path). Errors were rare in the forward condition and much more frequent in backward and rotated conditions. In addition, the time in ``off-path guidance'' was less in the ``Expected'' and the ``Preview”'' groups than the ``Un-Expected'' group, showing the benefits of the greater exploration during training (Fig.~\ref{fig:trainingarea}). The time devoted to exploration (red bars, bottom) also depended on the spatial relationships of the training and testing mazes, with the same pattern as for the other phases: least exploration in the forward condition and most in the rotated condition. 

The ``Preview'' group showed some noteworthy patterns. The ``Preview'' group showed less exploration in the testing trial (and less time spent off the path as well) than either ``Expected'' or ``Un-Expected'' groups. This means that when subjects scanned the maze, but were not allowed to move the mouse, they learned more. 

Notes that exploration reduced, but did not eliminate errors. This can be seen during training trials (Fig.~\ref{fig:trainingarea}), in which the greater exploration of the ``Expected'' groups did not eliminate errors, and in the testing trials (Fig. 9, off-path blue bars), in which off-path errors still occurred, even for the ``Preview'' group. Either the subjects did not explore enough, or there are limits to the benefits of explorations.

\section{General Discussion}

The current study aims to understand the underlying processes and strategies during a challenging visual-motor navigation task, maze solving. Maze solving makes demands on vision, attention and memory. By tracking eye and mouse movements, the strategies people use and how people learn can be inferred. The main findings can be divided into two main aspects: how eye-hand coordination links to the corresponding strategies and the learning effects. 

One of the main findings of this study is the emergence of two types of eye-hand coordination: guidance mode and exploration mode. These phases rarely overlapped. There was also a trade-off between the two. For example, people can either conduct more exploration in order to reduce the errors, which would be reflected as more ``exploration'' time and less ``off-path guidance'' time, or alternatively, explore less, which increases the errors, but minimizes the use of memory. The adjustment of strategies depended on the difficulty of mazes, the spatial relationships between training and testing mazes, expectations and previews. 

The results also showed that mazes can be learned quickly, and that learning did not require that the maze be traveled using the mouse. Learning was shown by the fact that the testing trials had shorter solving times than the training trials in both the forward and backward conditions. Learning was also evident in the rotated condition, when the mazes in the testing trials were hard to recognize, and where the learning led to poorer performance because the subjects tried to apply the paths expected from the training mazes to the testing mazes, where the configuration had changed due to rotation. All of this evidence shows that people could find and remember the paths with very little practice or exposure. 

These results have implication for strategies. There is a critical decision subjects need to make during solving mazes, namely, whether to try to keep moving as fast as possible in the maze at a risk of error, or stop moving and to rely on exploration with the eye to find the correct path. Errors (travel on the wrong path) are time-consuming because people have to back-track, while exploring would seem to be less time-consuming, since saccades are so fast. It would see that in order to achieve best performance in maze solving, people should use the eye to explore the mazes and not have any errors at all (not travel on wrong path). The data shows people did explore, and exploring was helpful in that more exploration found in the ``Expected'' and the ``Preview'' groups led to fewer errors in the testing trials (Figs.~\ref{fig:testingarea} and ~\ref{fig:meanbars}). However, people still spent some time on the wrong path. This suggests that people did not explore enough, or there are limits in the value of explorations. These limits were not due to the total absence of memory since the results showed that considerable learning occurred. But the cost in time or effort of using memory might prevent people to conduct more explorations. 

If people did not explore enough, and since we assume people are using a good strategy, an optimization computation must be involved. What are people optimizing? Previous work discussed several possible optimization strategies. For example, the optimization could be minimizing the use of the internal memory, and using the ``world'' as external memory (Ballard, et al., 1995). Epelboim and Suppes (2001) pointed out that people use their memory up to its limit and then use the eye to get more information from the display. The maze solving task is different from those studies previous because in this task, the cost of errors (travel on the wrong path) is time-consuming. Thus, failure to explore and use memory has more negative consequences than the tasks such as block-copying (Ballard, et al,. 1995), block-stacking (Flanagan \& Johansson, 2003), or problem-solving (Kong, et al., 2010). Given the fact that the ``Preview'' group performed much better than the ``Expected'' group, it suggests that either people prefer to avoid the use of memory, or there is a upper bound on the benefit of using memory during performance of a motor task, perhaps because of costs of retrieval. It is also possible that the main benefits of exploration are limited to long-term (across trial) benefits, instead of immediate benefits (within trial). This is supported by the fact that exploration was more effective at reducing errors in testing trials (Fig.~\ref{fig:testingarea}) than in training trials (Fig.~\ref{fig:trainingarea}). Thus, multiple types of memory maybe involved.

\section{Applications}

The present study raises new questions about how people devise optimal strategies for performing visual-motor tasks. These strategies involve trade-offs between memory and visual guidance and involve consideration of the cost in time or effort of making errors vs. the cost in time or effort of preventing errors. Unearthing the strategies of this type of trade-off can be valuable. With these strategies, we can understand how humans process information, including: (1) what should be stored in memory; (2) when to input new information; (3) how to update the states given the new inputs and the previous components in memory; and (4) how the updates change the decisions. A better understanding the mechanisms that determine the optimal strategies used in natural tasks could contribute to the AI and robotics fields. For example, with the current maze navigation task, we might be able to categorize people into different groups by examining the path planning patterns. Different types of human-machine interfaces could be provided to different groups.  For example, when using a GPS or other devices for guidance during road navigation, information about individual styles and capacities could determine how much information should be provided to its user, so that it would be just enough. Finally, understanding human strategies of trading off exploration and guidance can inform algorithms that aim to behave with human-like intelligence. This may be particularly useful for devising models to guide the robots efficiently on the basis of sensory and remembered evidence.  By allowing robots to use similar strategies as the humans, it may help facilitate human/robot communication, as well as the development of ways to use machine algorithms to compensate the limits of humans and maximize the benefits.


\end{document}